\documentclass[a4paper,conference]{IEEEtran}
\usepackage{cite}
\usepackage{algorithm}
\usepackage{multirow}
\usepackage{tabularx}
\usepackage{booktabs}
\usepackage{caption}
\usepackage{subfigure}
\ifCLASSINFOpdf
   \usepackage[pdftex]{graphicx}
   \graphicspath{{D:/Paper/LZQ/DA-refinenet/ICPR20\_DARefine/figures/pdf/}{D:/Paper/LZQ/DA-refinenet/ICPR20\_DARefine/figures/jpeg/}}
\else
   \usepackage[dvips]{graphicx}
\fi
\usepackage{amsmath}
\usepackage{algorithmic}

\usepackage{array}

\ifCLASSOPTIONcompsoc
  \usepackage[caption=false,font=normalsize,labelfont=sf,textfont=sf]{subfig}
\else
  \usepackage[caption=false,font=footnotesize]{subfig}
\fi
\usepackage{url}

\ifCLASSINFOpdf
\usepackage[pdftex]{graphicx}
\graphicspath{{D:/Paper/LZQ/DA-refinenet/IEEEconf\_ICPR18/}}
\else
\usepackage[dvips]{graphicx}
\graphicspath{{D:/Paper/LZQ/DA-refinenet/IEEEconf\_ICPR18/}}
\fi
\begin{document}
%
\title{DA-RefineNet: Dual-inputs Attention RefineNet for Whole Slide Image Segmentation}

\author{\IEEEauthorblockN{Ziqiang Li\IEEEauthorrefmark{1}, Rentuo Tao\IEEEauthorrefmark{1}, Qianrun Wu\IEEEauthorrefmark{2}, Bin Li\IEEEauthorrefmark{1}}
\IEEEauthorblockA{\IEEEauthorrefmark{1}CAS Key Laboratory of Technology in Geo-spatial Information Processing and Application Systems,\\ University of Science and Technology of China, Hefei, Anhui, China}
\IEEEauthorblockA{\IEEEauthorrefmark{2}School of foreign studies, Hefei University of Technology, Hefei, Anhui, China\\\{iceli, trtmelon\}@mail.ustc.edu.cn, tsianrun@gmail.com, binli@ustc.edu.cn}
}

%


\maketitle

\begin{abstract}
Automatic medical image segmentation has wide applications for disease diagnosing. However, it is much more challenging than natural optical image segmentation due to the high-resolution of medical images and the corresponding huge computation cost. The sliding window is a commonly used technique for whole slide image (WSI) segmentation, however, for these methods based on the sliding window, the main drawback is lacking global contextual information for supervision. In this paper, we propose a dual-inputs attention network (denoted as DA-RefineNet) for WSI segmentation, where both local fine-grained information and global coarse information can be efficiently utilized. Sufficient comparative experiments are conducted to evaluate the effectiveness of the proposed method, the results prove that the proposed method can achieve better performance on WSI segmentation compared to methods relying on single-input.
\end{abstract}

\renewcommand{\thefootnote}{}
\footnote{Accepted by ICPR2020}
\footnote{The source code of this work are available at: \url{https://github.com/iceli1007}.}
\footnote{Bin Li is the corresponding author}

%
\IEEEpeerreviewmaketitle

\section{Introduction}
We focus our attention on breast cancer pathological image segmentation in this paper. Studies\cite{mcguire2016world} have shown that early canceration screen and timely diagnosis can do great help for curing breast cancer, hence automatic pathological image segmentation and analyzing tools are meaningful for improving screening efficiency and mitigating the problem of limited medical resources.

In recent years, deep neural networks \cite{lecun1998gradient,szegedy2015going,szegedy2016rethinking} have outperformed the state of the art in many computer vision tasks like classification, detection, and semantic segmentation. For medical image segmentation, the common way is to slide a window on pathology images to get train slices, which can be used for training deep segmentation models\cite{ronneberger2015u,li2018h,dubost2017gp}. The U-Net\cite{ronneberger2015u} based deep segmentation methods and its variants achieve good performance on slice image segmentation but fail to utilize global information to supervise the segmentation process as human specialists do. Segmentation on pathological images is still challenging\cite{sirinukunwattana2018improving} for the problem of small receptive field caused by large image size.

 U-Net and its variations are the most dominant method used for semantic segmentation usually composed of two parts, an encoder and a decoder with skip-connections. These models adopt sliced medical images (local regions) as input thus no global information were used. For human specialists, they evaluate both global and local regions of medical images to derive an accurate diagnosing results. Inspired by the diagnosing process of human specialists, we proposed a dual-inputs (slice image and full image) model for whole slide breast cancer image segmentation. When it comes to whole slide images, comparing with single-input based models, such dual-inputs architecture can enlarge the receptive field and offer auxiliary global context information to the model.
\begin{figure*}
	\centering
	\includegraphics[width=0.8\textwidth]{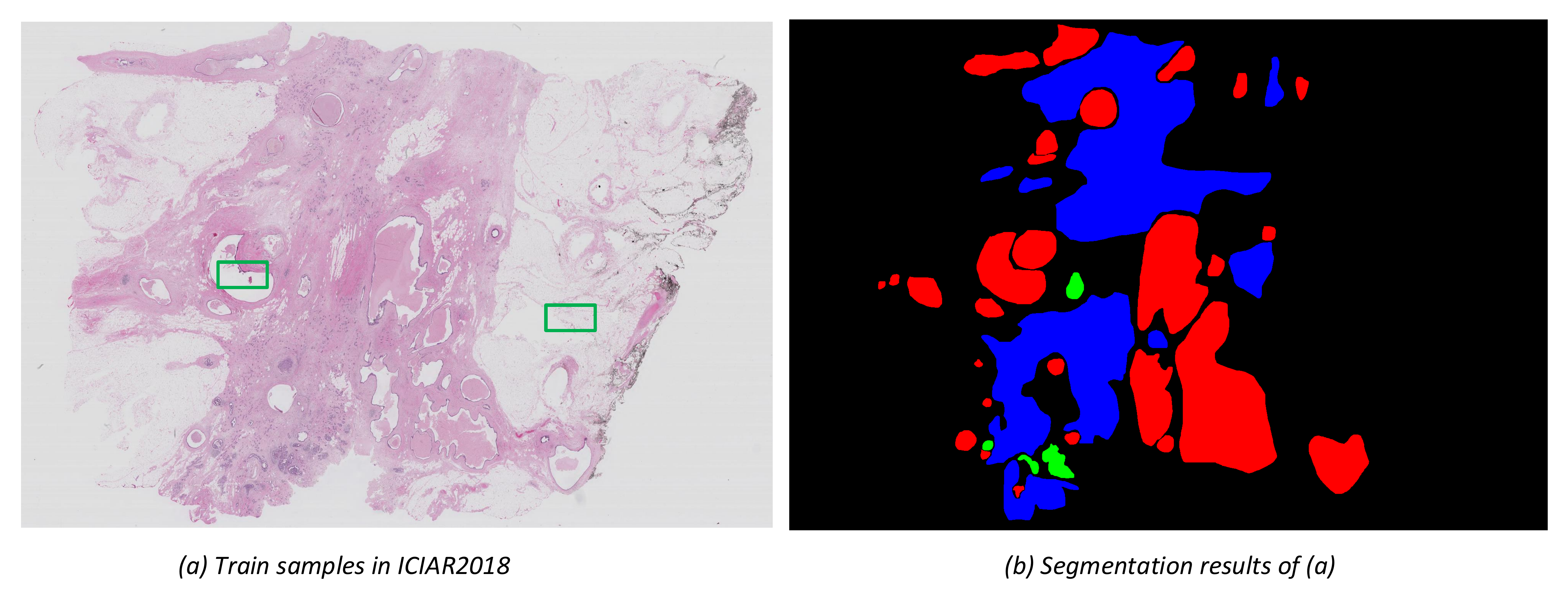}
	\caption{Demonstration of training medical images and its corresponding segmentation results. (a) training samples in ICIAR2018 dataset, where green rectangles denote sliced regions; (b) segmentation results of (a), color pixels represent different region types (normal, benign, in situ, invasive).}
	\label{fig:TrainSample}
\end{figure*}

\begin{figure*}
	\centering
	\includegraphics[width=0.85\textwidth]{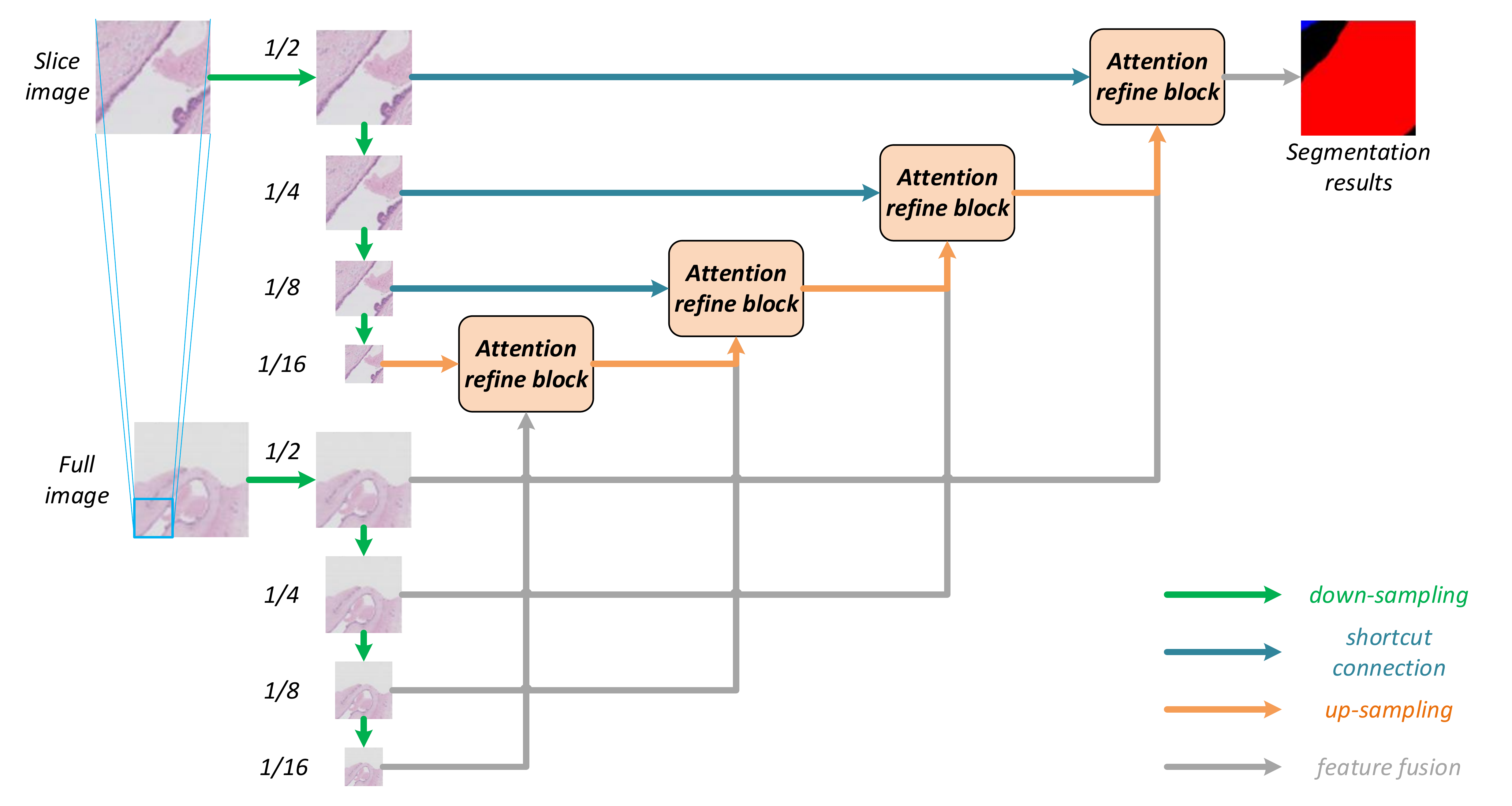}
	\caption{Model architecture of the proposed DA-Refinenet. Here slice image was sliced from the original full image and full image was resized (down-sampled 10 times to have the same size as slice image) before been forwarded to the model for processing. Slice image hold the fine-grained local information while full image hold coarse global context info.}
	\label{key:DA-RefineNet}
\end{figure*}

 It's easy to find that sliced patches with similar texture sometimes were labeled differently due to the context discrepancy, which can be seen in the left image of Fig. \ref{fig:TrainSample}. Intuitively, this may caused by the discrepancy between surrounding regions of different slice images. For a specific region, more informative context information can be derived by simply increasing the size of slide window, however, it will also bring more computation costs. In order to balance slice image size and global context information, we propose a dual-inputs attention network named DA-Refinenet to combine fine texture features and coarse spatial features together, which allows a larger receptive field with little computation cost increasing. In the proposed model (Fig. \ref{key:DA-RefineNet}), full image is down-sampled to the same dimension with slice image to reduce the computation cost. The down-sampled full image can give enough auxiliary semantic information hence we can get larger receptive field under limited memory. Besides, although the proposed method is designed for breast caner image segmentation, it can also be extended to other WSI processing tasks. The main contributions of this paper can be summarized as below:
\begin{itemize}
	\item  We proposed a new attention based dual-inputs framework for whole slide image segmentation, which can incorporate global image information and obtain larger receptive fields. It proves that the proposed method can achieve better performance compared to methods that rely on single-input.
	\item  We explore several feature fusion strategies and propose a simple but effective fusion strategy based on attention mechanism under the intuition that coarse global features can help reorganize fine-grained local features.
\end{itemize}

The paper is organized as follows: in Section II, we give the related works introduction; in section III, detailed information about the proposed method will be presented; then is the implementation details and experiment results analysis.

\section{Related works and Prerequisite}
Traditional methods usually adopt 'thresholding'\cite{fan2001automatic}, 'region growing'\cite{kaya2003new}, 'classifier' or 'clustering'\cite{altunbay2009color,liu2007survey} etc. for medical image segmentation. Along with the success of deep learning techniques in visual tasks, segmentation models which adopt deep architecture also outperform these traditional methods greatly.

Long et al.\cite{long2015fully} first proposed a fully convolutional network for pixel-wise image segmentation. Ronneberger et al.\cite{ronneberger2015u} adopted the encoder-decoder architecture with skip-connections for medical image segmentation, which was proved effective on multi-scale feature fusing. Lin et al.\cite{lin2017refinenet} proposed a new module RefineNet, which was inspired by the idea of residual connection\cite{he2016deep}. It can make dense prediction more accurate and capture the background context information more efficiently by utilizing information lost over down-sampling operations and chained residual pooling. Yu et al.\cite{yu2018learning} proposed a feature discrimination network for solving inter-class indistinction and intra-class inconsistency problems in semantic segmentation. The above-mentioned works also proposed several feature fusion modules to improve the ability of combining high-level semantic information and low-level structural information.

For WSI segmentation tasks, sliding window is the most commonly used method for splitting large medical image into multiple slice patches. Methods for segmenting WSI can be mainly divided into two categories: classifier based patch-wise segmentation and end-to-end pixel-wise segmentation. The first kind of methods treat segmentation as classification problems and utilize classifier to make predictions on sliced image patches. Cruz-Roa et al.\cite{cruz2017accurate} proposed to use a classification network for accurate invasive breast cancer detection in WSI. Hou et al.\cite{hou2016patch} proposed to adaptively combine patch-level classification results by an EM-algorithm based post-processing. Korsuk et al.\cite{sirinukunwattana2018improving} also utilized multi-scale information in classifier training for deriving better segmentation accuracy. However, the above mentioned methods, which give each patch the same label, are easy to generate contradictive segmentation results on patch borders.

The other kind of methods for pixel-wise whole side images segmentation were usually trained in an end-to-end\cite{guo2019fast} manner. Cruz-Roa et al.\cite{cruz2014automatic} first propose to use deep learning techniques for breast cancer whole-slide segmentation task and achieved better performance than manual designed features. Gu et al.\cite{gu2018multi} proposed a FCN-based multi-resolution network for WSI segmentation. Similarly, these methods also does not offer any qualitative analysis of features at different scales. Moreover, the feature fusion strategy in medical segmentation networks also left unexplored. Tokunaga et al.\cite{Tokunaga_2019_CVPR} proposed an adaptive weighting multi-field-of-view CNN semantic segmentation network for whole slide images, which ensemble several expert CNNs for images of different magnifications by adaptively changing the weight of each expert according to input images. This idea shares some similarity with ours, the different is that we adopted the attention based dual-inputs component, not just simple feature weighting as\cite{Tokunaga_2019_CVPR}. Moreover, since the feature weighting vector proposed in this article is generated by a neural network, which also increase the model parameters.

In this paper, we proposed an end-to-end model based on Refinenet\cite{lin2017refinenet} for breast cancer pathological image segmentation. The main difference between Refinenet and Unet\cite{ronneberger2015u} lies in the usage of a unique feature fusion block: "Refine Block", which can be divided into three parts: 
\begin{itemize}
	\item  Residual Convolution Unit (RCU); A convolutional module based on residual connection, where BN layer was removed compared to original Resnet\cite{he2016deep}.
	\item  Multi-size fusion; Methods applied in up-sampling and feature fusion operations to keep input and output the same size for medical image segmenation.
	\item Chain residual pooling(CRP); Convolution pooling operations for efficiently fusing features of different sizes. Receptive field can be expanded through this chained pooling operation.
\end{itemize}

\section{Dual-input Attention RefineNet}
\subsection{Problem Statement}
Semantic medical image segmentation aimed at precisely classify each pixel or region to correct pathology types. The objective of a segmentation model $f_{\theta}$ can be formulated as below:
\begin{align}
\begin{split}
&\{X^i,M^i\},\,i=1,\cdots,N\\
&\min_{\theta} f_{\theta}=\sum_{i=1}^{N}D\left(f_{\theta}\left(X^i\right),M^i\right)
\end{split}
\end{align}
where $N$ and $\theta$ are the number of training samples and model parameters respectively, $X$ and $M$ are train images and its corresponding label mask, D is the distance metric. For large medical images, each train image and label mask $\{X,M\}$ were first been split into $T$ patches $\{X_i,M_i\}, i=1,\cdots,T$ before been forwarded to the segmentation model $f_{\theta}$. As for even larger whole slide images, each patch $\{X_i,M_i\}$ were further split into K smaller slices $\{{X_S}_i^j,{M_S}_i^j\},j=1,\cdots,K$ for processing. Below are the objective function of common single-input whole slide image segmentation methods:
\begin{align}
\begin{split}
&\{{X_S}_i^j,{M_S}_i^j\},\,i=1,\cdots,T.\,\,j=1,\cdots,K\\
&\min_{\theta} f_{\theta}=\sum_{i=1}^{T}\sum_{j=1}^{K}D\left(f_{\theta}\left({X_S}_i\right),{M_S}_i\right)
\end{split}
\end{align}
where ${X_S}_i^j$ and ${M_S}_i^j$ are smaller slices and label masks derived from image patch $X_i$ and $M_i$. In this paper, we propose a dual-input attention network for WSI segmentation, where we use $X_i$ as an auxiliary input for ${X_S}_i^j$ segmentation. The process can be formulated as below:
\begin{equation}
\min_{\theta} f_{\theta}=\sum_{i=1}^{T}\sum_{j=1}^{K}D\left(f_{\theta}\left({X_S}_i^j,X_i\right),{M_S}_i\right)
\end{equation}

In the later part of this paper, we will denote ${X_s}_i^j$ and $X_i$ as slice image and full image respectively, the auxiliary full image input can provide surrounding context information and enlarge the receptive field for slice image segmentation.

\subsection{Dual-input Attention RefineNet}
The architecture of proposed DA-RefineNet was demonstrated in Fig. \ref{key:DA-RefineNet}, where we can see the model mainly composed of three parts: two encoders ($ENC_{slice}, ENC_{full}$) for slice image and full image processing, and a refine decoder ($DEC_{refine}$) for producing segmentation results. Color arrows in green, blue, orange and gray represent down-sampling, skip-connection, up-sampling and feature fusion operations respectively.

We adopt the encoder-decoder structure as U-Net in the proposed model and combine coarse global context info with fine-grained local details together to increase the receptive field through attention refine blocks. $ENC_{slice}$ and $ENC_{full}$ can encode slice image ${X_S}_i^j$ and full image $X_i$ into different scales respectively and obtain corresponding high-level semantic features. The encoding process of $ENC_{slice}$ and $ENC_{full}$ can be formulated as below:
\begin{align}
\begin{split}
h_S^1,h_S^2,h_S^3,h_S^4&=ENC_{slice}\left({X_S}_i^j\right)\\
h_F^1,h_F^2,h_F^3,h_F^4&=ENC_{full}\left(X_i\right)
\end{split}
\end{align}
where $h_S$ and $h_F$ represent encoded hidden features at different scale of slice image and full image respectively. For easy explanation, we will obliterate the subscript and use $X$ and $X_s$ to denote full image and slice image respectively in the below contents. The decoding process of DA-RefineNet can also be formulated as below:
\begin{align}
\begin{split}
o_i&=DEC_{refine}\left(h_S^i,h_F^i\right),\,i=1\\
o_i&=DEC_{refine}\left(h_S^i,h_F^i,o_{i-1}\right),\,i=2,3,4
\end{split}
\end{align}
where $o_i$ denote the output of i-th attention refine block, $o_4$ is the final segmentation result.

\begin{figure}[h]
	\centering
	\includegraphics[width=0.5\textwidth]{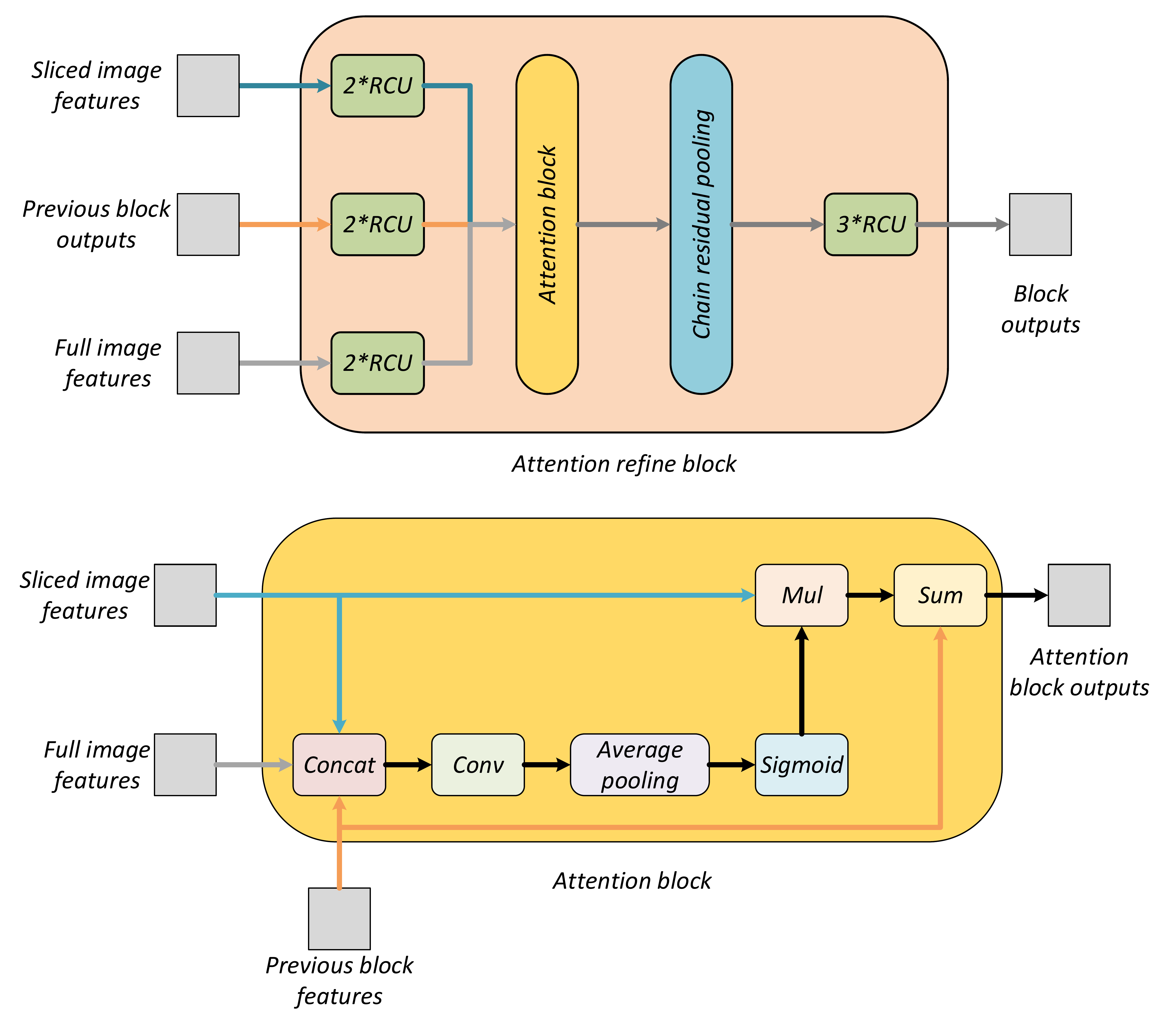}
	\caption{Attn-Refine Block: the top part denote the architecture of a attention refine block while the bottom part denote the attention block structure.}
	\label{key:Attn-Refine}
\end{figure}

Based on the intuition that global coarse images along with fine-grained local images features can help improve model performance, we adopt attention mechanism and proposed the attention refinement block(Attn-Refine). Detailed architecture of the proposed Attn-Refine block can be seen in the top part of Fig. \ref{key:Attn-Refine}, where color arrows represent different type of input features. Attn-Refine block contains three individual components: Attention block, RCU (residual convolution unit) and CRP (chained residual pooling), where the latter two module have been introduced in Section II. The proposed attention block is designed for feature fusion, which can be seen in the bottom part of Fig. \ref{key:Attn-Refine}. To make comparison, we keep RCU and CRP module the same as RefineNet\cite{lin2017refinenet}.

For each input path, features are passed sequentially through two residual convolution units (RCU), which is a simplified version of original ResNet  convolution unit, but with batch-normalization layers removed. Then the residual convolution features of different inputs were aggregated in the attention block, whose concrete architecture can be seen in the bottom part of Figure. \ref{key:Attn-Refine}. The input of the k-th Attn-Refine block are the features of sliced image $x^j_i$, previous block output $O_{k-1}$, and full image $x_i$, thus attention refined features can be derived through the below equation:
\begin{equation}
X_{fusion}=fusion\left(X_S,X,O_{k-1}\right)
\end{equation}
\begin{equation}
W_{attn}=\mathbf{W}(X_{fusion})
\end{equation}
\begin{equation}
O_k=W_{attn}*X_S+O_{k-1}
\end{equation}
where $X_{fusion}$, $W_{attn}$ and $O_k$ represent the fused feature, attention weights and output of k-th attention block respectively.

CRP\cite{lin2017refinenet} was built as a chain of multiple pooling blocks, each consisting of one max-pooling layer and one convolution layer. It was able to re-use the result from the previous pooling operation and thus access the features from a large region without using a large pooling window. Moreover, full image features are only been concatenated as auxiliary information during the feature fusion process in the proposed DA-RefineNet. It is proved very effective to incorporate large scale coarse features through the above mentioned feature fusion scheme.
\begin{figure}[h]
	\centering
	\includegraphics[width=0.5\textwidth]{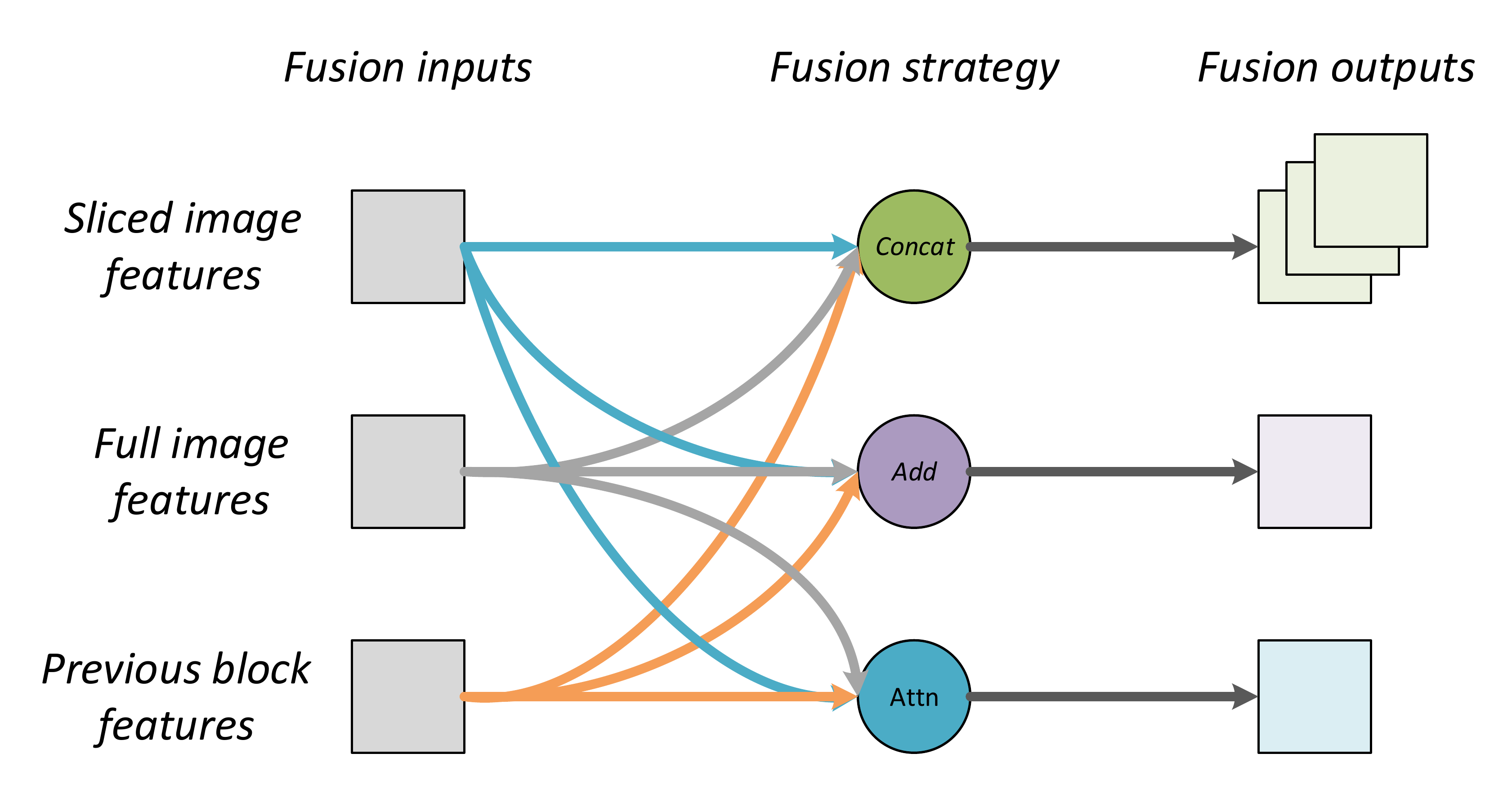}
	\caption{ Three methods of feature fusion.  }
	\label{key:Fusion-Strat}
\end{figure}

Besides, in order to explore the relationship between several type of input features, we also make comparisons between the proposed attention fusion strategy and other commonly used 'concat' and 'add'. The illustration of three feature fusion strategies can be seen in Fig. \ref{key:Fusion-Strat}.
\begin{itemize}
	\item Concatenation fusion: 'concat' fuse different features together by simply concatenating them along the channel dimension, thus increased number of feature channels will be derived. Moreover, the concatenated features are contributed equally, which is different with 'attention'.
	\item Add fusion: Direct addition of corresponding channels of different type input features. This method has the lowest computational complexity, however the relationship among channels is destroyed during the addition process, so there exist information loss through this operation.
	\item Attention fusion: It provide a way for automatically learning attention weights for feature fusion, which is more reasonable than 'concat' or 'add'. We use the coarse global image as an auxiliary information to promote the reorganization of fine local features and use attention for feature fusion in the attn-refine block.
\end{itemize}

\subsection{Evaluation Metrics }
In order to evaluate the performance of the proposed method, we follow previous works \cite{marami2018ensemble}\cite{galal2018candy} and choose 'MIoU' (mean IoU), 'Accuracy' and 'Score' as the quantitative evaluation metrics.

MIoU is the most commonly used metric in semantic segmentation, which calculates IoU scores for each class and average them across all categories:
\begin{equation}
IoU=\frac{DR\cap GT}{DR\cup GT},\,MIoU=\frac{\sum IoU_i}{N}
\end{equation}
where DR, GT and N denote as segmentation result, ground truth and number of categories respectively.

'Accuracy' was usually calculated by quantify the ration of correctly classified pixels. 'Score' is a dedicated metric for this task, which can be calculated by the below equation.
\begin{align}
\begin{split}
h=\sum\limits_{i=1}^N\max&(|gt_i-0|,|gt_i-3|)*\\
&\left[1-\left(1-pred_{i,bin}\right)\left(1-gt_{i,bin}\right)\right]
\end{split}
\end{align}
\begin{equation}	
score=1-\sum\limits_{i=1}^N|pred_i-gt_i|/h
\end{equation}
where "pred" is the output predictions on categories (0, 1, 2, 3), "gt" represent the ground truth, and the subscript bin indicate binarized results. 'Score' is based on accuracy but is designed to penalize more pixels away from real values. It need to notice that true negative cases for prediction and ground truth are both 0 (normal class) are not counted in the denominator.

\section{Implementation Details}
\subsection{ICIAR2018 Dataset}
ICIAR2018\cite{aresta2019bach} dataset is composed of three type of images: hematoxylin and eosin (H\&E), stained breast histology microscopy and whole-slide images (WSI) respectively. It encompassed a total of 400 microscopy images which were labeled as normal, benign, in situ carcinoma or invasive carcinoma according to the predominant cancer type by two medical experts. The dataset also contains 10 whole slide images, which are high resolution images that contain entire sampled tissue. Multiple region types may exist in a single WSI. Another thing need to mention is that we have not use the microscopy images for pre-training in this work, which means we only use whole slide images and its sliced patches for training. The train set consists of 3k patches sliced from whole slide images 2$\sim$4 and 6$\sim$9. In order to ensure data balance, we select a total of 2k patches containing benign or in situ and use data augmentation techniques such as random flip, random crop for data preparation. The validation set consists of the 500 relatively balanced patches selected in image 10 and the test set consists of a total of 3k patches of all the patches in image 5. Since that there also exist some normal and invasive samples in these patches, thus we randomly choose another 1k patches without benign and in situ regions.

\subsection{Model Architecture and Hyper-parameter settings}
In this work we use negative log likelihood (NLL) as the loss criteria for the proposed model. NLL loss is also called cross-entropy loss, which can be formulated as the below equation, where t is a 4-dimension one-hot vector and y is the softmax output probabilities for normal, benign, in situ and invasive respectively.
\begin{equation}
\label{eq:alpha}	
NLL_{loss2d}(t,y)=-\sum\limits_{i}t_i\log y_i
\end{equation}

The slice image encoder and full image encoder network of the proposed DA-RefineNet was build on ResNet-152 and ResNet-50 respectively, the encoder was composed of four attention refine blcok. Other comparative model network architectures can be found at Table \ref{table:dual_single} and Table \ref{table:fusion_compare}. All the experiments were conducted under Pytorch framework. The model were all trained by SGD\cite{bottou2010large} optimizer and batch size was set as 12. Nowadays, strong post-morphological processing techniques are often adopted to optimize the segmentation results. However, the post-processed results cannot reflect the true performance or defects of the method, thus we do not apply any processing or post-processing techniques in the experiments.
\begin{table*}
	\renewcommand\arraystretch{1.5}
	\caption{Quantitative Comparison of Dual-inputs vs Single-input}
	\label{table:dual_single}
	\setlength{\tabcolsep}{2mm}
	\centering
	\begin{tabular}{c | c | c | c c c c c c c c }
		\toprule
		{\bfseries Encoder}& {\bfseries Decoder}& {\bfseries Input}& {\bfseries $MIoU$}& {\bfseries $IoU_0$}& {\bfseries $IoU_1$}& {\bfseries $IoU_2$}& {\bfseries $IoU_3$}&  {\bfseries $Accuracy$}& {\bfseries $Score$}& {\bfseries $Params$} \\
		\midrule
		\multirow{1}{*}{U-net}&
		U-net& Single-input&31.2& 45.5 &25.0 &20.1 &50.2&  57.2& 49.1& 150M\\
		\midrule
		\multirow{3}{*}{Resnet-50}&
		\multirow{3}{*}{Add-Refine}&
		Single-input&36.2& 60.5& 28.0& 21.8& 55.0&  67.1& 58.8& 334M\\
		&& Multi-size dual-inputs& 44.8& {\bfseries 60.9}& 24.2& 31.1& {\bfseries 63.1}&  74.1& 71.2& 441 M\\
		& &{\bfseries Our dual-inputs}&{\bfseries45.9}& 55.3& {\bfseries 32.9}& {\bfseries 37.1}& 58.6& {\bfseries 75.1}&{\bfseries 71.1}& 441M\\
		
		\hline
		\multirow{3}{*}{Resnet-101 }&
		\multirow{3}{*}{Add-Refine}&
		Single-input&42.2& 58.5& 22.3& 29.8& 58.1&  73.8& 69.0& 417M\\
		& &Multi-size dual-inputs&44.4& {\bfseries 59.1}&
		28.4& 33.0& {\bfseries 62.9}&  73.7& 69.9& 517M\\
		& &{\bfseries Our dual-inputs}& {\bfseries46.5}& 58.0& {\bfseries 34.5}& {\bfseries 38.9}& 61.4 &{\bfseries75.1}& {\bfseries 71.6}& 517M\\
		
		\hline
		\multirow{3}{*}{Resnet-152}&
		\multirow{3}{*}{Add-Refine}&
		Single-input&39.7& 59.0& 27.7& 28.3& 52.5&  72.7& 69.7& 480M\\
		& &Multi-size dual-inputs& 44.7&{\bfseries 60.4}& 25.3& 31.2& {\bfseries 61.8}&  74.9& 71.1&580M\\
		& &{\bfseries Our dual-inputs}&{\bfseries 46.8}&59.5&{\bfseries 38.1}& {\bfseries 28.8}& 60.9&  {\bfseries 75.5}&{\bfseries 71.5}& 580M\\
		\bottomrule
	\end{tabular}
\end{table*}

\begin{table*}
	\renewcommand\arraystretch{1.5}
	\caption{Quantitative Comparison of Feature Fusion Strategies}
	\label{table:fusion_compare}
	\setlength{\tabcolsep}{3mm}
	\centering
	\begin{tabular}{c | c | c  c c c c c c }	
		\toprule
		{\bfseries Dual-inputs Encoder}& {\bfseries Fusion strategy}& $MIoU$&	$IoU_0$& $IoU_1$& $IoU_2$& $IoU_3$& $Accuracy$& $Score$\\
		\midrule
		\specialrule{0em}{1pt}{1pt}
		\multirow{3}{*}{ResNet50\_50 }
		& Concat&48.9&	60.1&	38.9&	39.2&	58.3&		75.7&	71.5\\
		
		& Add&	45.9&55.3&	32.9&	37.1&	58.6&		75.1&	71.1\\
		& {\bfseries Attention}&{\bfseries 51.0}&	{\bfseries 61.1}&	{\bfseries 40.9}&	{\bfseries 39.4}&	{\bfseries 62.3}&		{\bfseries 76.4}&	{\bfseries 72.0}\\
		\hline
		\multirow{3}{*}{ResNet101\_50}
		& Concat&47.8&	58.3&	36.6&	42.0&	60.0&		74.4&	71.9\\
		& Add&46.5&	58.0&	34.5&	38.9&	{\bfseries 61.4}&		{\bfseries 75.1}&	71.6\\
		& {\bfseries Attention}&{\bfseries 49.9}&	{\bfseries 59.3}&	{\bfseries 36.6}&	{\bfseries 44.0}&	59.7&		74.8&	{\bfseries 72.1}\\
		\bottomrule
	\end{tabular}
\end{table*}
\section{Experiments}
\subsection{Dual-inputs vs Single-input}
To evaluate the effectiveness of the proposed DA-RefineNet model, we first make comparison between segmentation models with single-input, and attention dual-inputs to demonstrate the superiority of the proposed attention dual-inputs architecture. Moreover, in order to prove that the performance improvement is derived from the proposed dual inputs architecture other than simple increase of the parameter amount, we also compare the proposed method with the multi-size dual-inputs (full image in multi-size dual-inputs models denote the resized version of the slice image with no content change).

For single-input models, we choose U-Net and three RefineNet variants (RefineNet-50, -101 and -152) for comparison. For multi-size dual-inputs and attention based dual-inputs, we choose three architecture variants (ResNet-50, -101 and -152) for slice image encoder and ResNet-50 as the full image encoder. The decoder for all the comparison model variants except U-Net are add-refine based decoders.

The comparison results can be seen in Table \ref{table:dual_single}, where $IoU_0 \sim IoU_3$ represent IoU scores of four categories (normal, benign, in situ and invasive cancer) respectively. MIoU represents the average IoU score across the four categories. Form the comparison results we can arrive the following conclusions:
\begin{itemize}
	\item Comparing with single-input methods, the proposed method achieve great improvement on $MIoU$, $Accuracy$ and $Score$. Especially, the proposed attention dual-inputs model improved $MIoU$ by 28\%, 10\%, 18\% under three encoder variants (ResNet-50, -101, -152) respectively compared with corresponding single-input method.
	\item Comparing with multi-size dual-inputs models, DA-RefineNet can achieve better performance with the same model size. We also find the proposed model (ResNet50\_50) can achieve better performance with fewer parameters comparing with single-input method RefineNet-152, which proved that the proposed attention dual-inputs model is more efficient and can achieve similar results with shallower model depth.
	\item Segmentation accuracy of 'benign' and 'in situ' are relatively low, which may have some relation with the class imbalance of our dataset. Although we have adopted some techniques to balance different classes, it is inevitable that there are more normal and invasive cancer regions than the other two types.
\end{itemize}

\subsection{Feature Fusion Strategies}
Intuitively, coarse global features can be adopted as auxiliary information for fine local features, which can also be viewed as feature fusion. We design two comparative experiments to evaluate the performance of different feature fusion strategies ('concat', 'add' and 'attention'). For each feature fusion strategy, we select two encoder variants (ResNet-50, -101) for slice encoder and ResNet-50 for full image encoder. Comparative results are shown in Table \ref{table:fusion_compare}, from which we can clearly see that 'attention' achieved best performance compared with the other two feature fusion strategy.

Moreover, a good feature fusion strategy can improve the model robustness with respect to the model depth to some extent. The feature extraction capacity of ResNet-50 is lower than that of ResNet-101, similar conclusion can be arrived by observing the results of 'concat' and 'add' fusion strategy. However, for the attention-based feature fusion strategy, the result of ResNet50 is better than ResNet-101, which means we can accelerate the convergence and get better segmentation results by adopting attention-based feature fusion. This can reduces the computation cost and provides a practical way for real-time segmentation.
\begin{figure*}
	\centering
	\includegraphics[width=\textwidth]{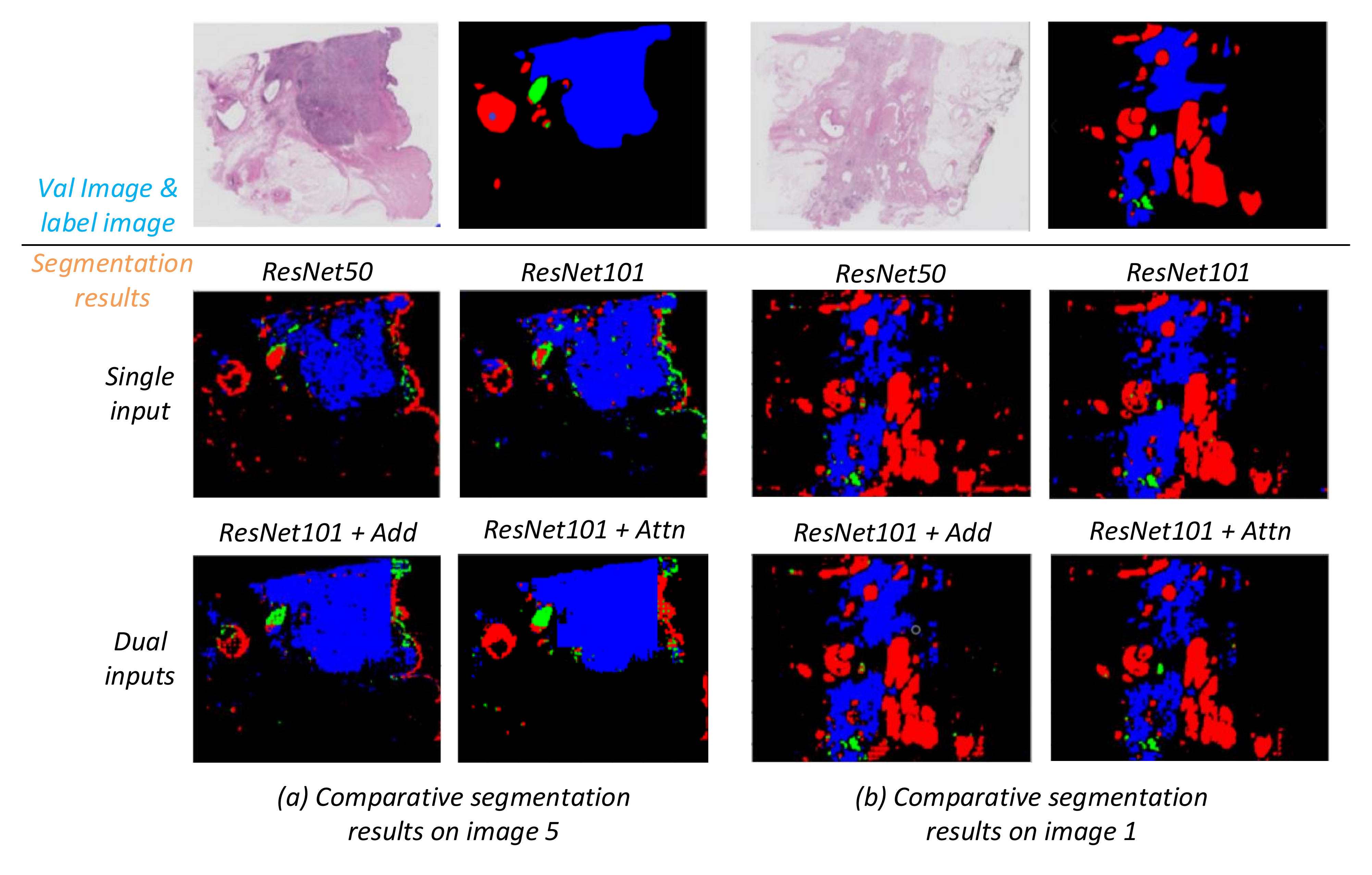}
	\caption{Visual segmentation results of image 5(left) and image 1(right). The top row represent the original image and its corresponding label mask respectively. The second rows give the segmentation results of single input with ResNet-50 and ResNet-101 respectively. The third row give the segmentation results of dual inputs of Resnet101-50 with feature fusion scheme "Add" and "Attention" respectively.}
	\vspace{-0.2cm}
	\label{fig:compare_fig}
\end{figure*}
\subsection{Qualitative Evaluation Results}
The visual segmentation results were demonstrated in Fig. \ref{fig:compare_fig} for better illustration of the effectiveness of the proposed method, where the top row denote validation images and its corresponding label masks, the bottom parts denote segmentation results of single-input and proposed attention based dual-inputs methods of two validate images.

By observing the segmentation results of single-input methods we can see that there exist a lot of red noise on the black background, which means the single-input methods tend to classify normal regions as benign, which is consistent with the phenomenon we mentioned in the motivation at the beginning of this paper. Moreover, due to the lack of global information, some parts of the training data with similar textures and normal areas are likely to be labeled as benign, which can mislead the network, causing the network to be inferior for benign and normal and result in mis-classification. 

Compared with single-input and multi-size dual-inputs methods, the segmentation results based on attention dual-inputs model (third row of Fig. \ref{fig:compare_fig}) have a relatively clean background. We also compare the feature fusion strategy based on 'add' and 'attention'. Comparing to the simple addition of two features, we find that the proposed method of adopting global coarse semantic information to help local fine features reorganization achieved better results. There exist some blur and noise for the reason that we did not apply any post-processing techniques. However, these results can reflect some weakness of the proposed method in whole slide image segmentation, which may be improved in future works.

\section{CONCLUSION}
In this paper, we proposed a new dual-inputs framework DA-RefineNet based on attention for whole-slide breast image semantic segmentation. The idea of adding full image as auxiliary information was inspired by the diagnosing process of human specialists. The qualitative and quantitative comparative experiment results both proved the superiority of the proposed attention based dual-input model compared to single-input and multi-size dual-inputs models. Moreover, we also conduct experiments to evaluate the effect of several feature fusion strategies and the results indicate that the attention-based feature fusion strategy was superior to simple 'concat' or 'add' fusion operations. It indicate that coarse global information can promote fine local features reorganization and improve the network convergence and representative ability. We can use shallower feature extraction network to get better results, which indicate that model performance is not only dependent on the depth of feature extractors, but also correct prior knowledge and effective feature fusion strategies. The proposed method can also give insight and provide a general framework for future WSI segmentation works.

%
\section*{Acknowledgment}
This work was supported by the National
Natural Science Foundation of China under grand No.U19B2044 and No.61836011. We also want to thank the data provider organizer of the ICIAR2018 Grand Challenge.





%
\ifCLASSOPTIONcaptionsoff
\newpage
\fi

\bibliographystyle{IEEEtran}
\bibliography{bare_conf}

\end{document}